\documentclass[conference]{IEEEtran}

\usepackage{graphicx}
\usepackage{amsmath}
\usepackage{amssymb}
\usepackage{booktabs}
\usepackage{multirow}
\usepackage{hyperref}
\usepackage{url}
\usepackage{cite}
\usepackage{tikz}
\usetikzlibrary{positioning,arrows.meta,shapes.multipart}
\usetikzlibrary{shapes.geometric}
\usepackage{graphicx}
\usepackage{subcaption}

\title{AMP2026: A Multi-Platform Marine Robotics Dataset for Tracking and Mapping}

\author{
Edwin Meriaux$^{1,2,3}$ \quad
Shuo Wen$^{1,2}$ \quad
David Widhalm$^{4}$ \quad
Zhizun Wang$^{1,2}$ \quad \\
Junming Shi$^{1,2}$ \quad

Mariana Sosa Guzmán$^{1,2}$ \quad
Kalvik Jakkala$^{5}$ \quad
Bennett Carley$^{5}$ \quad
Elias Sokolova$^{5}$ \quad \\

Yogesh Girdhar$^{6}$ \quad
Monika Roznere$^{7}$ \quad
Jason O'Kane$^{5}$ \quad
Junaed Sattar$^{4}$ \quad
Gregory Dudek$^{1,2}$%
\thanks{$^{1}$ McGill University, Montreal, Canada. $^{2}$ MILA - Quebec AI Institute. $^{3}$ Université Paris-Saclay, CentraleSupélec. $^{4}$ University of Minnesota. $^{5}$ Texas A\&M University. $^{6}$ Woods Hole Oceanographic Institution. $^{7}$ Binghamton University.}
}
\IEEEoverridecommandlockouts

\begin{document}

\newcommand{\gpsicon}{%
\tikz[baseline=-2.2ex]{
    \node[
        diamond,
        fill=green!70!black,
        draw=green!50!black,
        inner sep=0pt,
        minimum size=4.5mm
    ] {};
}
}

\tikzset{
    title/.style={
        font=\bfseries\LARGE,
        align=center
    },
    cat/.style={
        font=\bfseries\Large,
        align=center
    }
}

\maketitle

\begin{abstract}
Marine environments present significant challenges for perception and autonomy due to dynamic surfaces, limited visibility, and complex interactions between aerial, surface, and submerged sensing modalities. This paper introduces the Aerial–Marine Perception Dataset (AMP2026), a multi-platform marine robotics dataset collected across multiple field deployments designed to support research in two primary areas: multi-view tracking and marine environment mapping. The dataset includes synchronized data from aerial drones, boat-mounted cameras, and submerged robotic platforms, along with associated localization and telemetry information. The goal of this work is to provide a publicly available dataset enabling research in marine perception and multi-robot observation scenarios. This paper describes the data collection methodology, sensor configurations, dataset organization, and intended research tasks supported by the dataset.
\end{abstract}

\begin{IEEEkeywords}
Marine robotics, multi-platform perception, visual tracking, marine mapping,
underwater vision, aerial–marine sensing, multi-view perception, robotics datasets
\end{IEEEkeywords}

\section{Introduction}

\begin{figure*}[t]
\centering
\begin{tikzpicture}[
    font=\small,
    arrow/.style={-{Stealth[length=2.3mm]}, thick},
    title/.style={font=\bfseries\large, align=center},
    cat/.style={font=\bfseries, align=center},
    item/.style={
        draw, rounded corners, thick,
        align=left, inner sep=4.5pt,
        text width=0.30\linewidth,
        minimum height=10mm
    },
    qbc/.style={fill=blue!12, draw=blue!55!black},
    bbd/.style={fill=orange!14, draw=orange!60!black}
]

\node[title, font=\bfseries\LARGE] (top) {Dataset};

\node[cat, font=\bfseries\Large, below=11mm of top, xshift=-0.25\linewidth] (trackH) {Tracking};
\node[cat, font=\bfseries\Large, below=11mm of top, xshift=0.25\linewidth] (mapH) {Mapping};

\draw[arrow] (top.south) -- ++(0,-5mm) -| (trackH.north);
\draw[arrow] (top.south) -- ++(0,-5mm) -| (mapH.north);

\node[item, qbc, below=5mm of trackH] (t1)
{\gpsicon Multi-Drone Tracking of Submerged Robotic Platforms};

\node[item, bbd, below=3.2mm of t1] (t2)
{\gpsicon Single-Drone Tracking of Surface Vessel};

\node[item, bbd, below=3.2mm of t2] (t3)
{Multi-Drone Tracking of Swimmers and Submerged Robots};

\node[item, bbd, below=3.2mm of t3] (t4)
{Single-Diver Tracking of Submerged Robots and Divers};

\node[item, bbd, below=3.2mm of t4] (t5)
{Single-Submerged Robot Tracking of Divers};

\node[item, bbd, below=5mm of mapH] (m1)
{Single-Drone Multi-Area Flights};

\node[item, bbd, below=3.2mm of m1] (m2)
{Drone--Surface Vessel Same-Area Observations};

\node[draw, rounded corners, align=left, inner sep=4pt,
      below=20mm of m2, font=\small] (leg) {%
\textbf{Legend:}\;
\tikz[baseline=-2.6ex]{
    \node[item, qbc, text width=12mm, minimum height=5mm, inner sep=1pt] (box) {};
}~Quebec \quad
\tikz[baseline=-2.6ex]{
    \node[item, bbd, text width=12mm, minimum height=5mm, inner sep=1pt] (box) {};
}~Barbados
\\[3pt]
\hspace*{2.8em}\gpsicon~GNSS Ground Truth Available
};

\end{tikzpicture}
\caption{AMP2026 dataset overview organized by task category. Colored blocks correspond to where the data of a particular class was collected, while the green symbol indicates the availability of GNSS ground truth.}
\label{fig:dataset_overview_blocks}
\end{figure*}
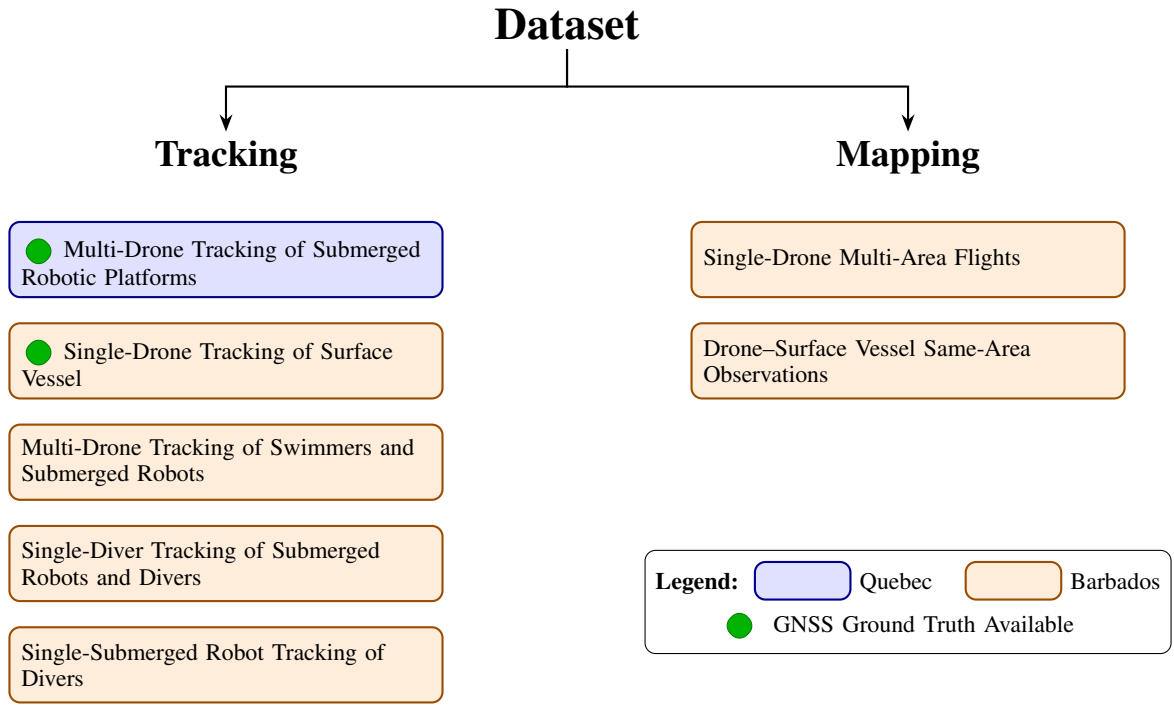

Marine environments present significant challenges for vision-based perception due to dynamic water surfaces, refraction at the air--water interface, specular reflections, turbidity, and rapidly changing illumination conditions. In particular, dynamic caustic patterns projected onto the seabed create high-frequency visual noise that distracts feature detectors, while the non-rigid motion of the water column itself violates the static-world assumptions common in terrestrial SLAM and tracking algorithms. These effects introduce geometric distortions and appearance variability that complicate visual tracking and map construction. In shallow-water environments, additional challenges arise from partial submergence of targets, surface motion, and viewpoint-dependent visibility changes, making consistent perception across time and viewpoints difficult~\cite{akkaynak2019sea,li2017watergan,islam2020fast}.

Recent progress in computer vision and robotics has been supported by publicly and privately available datasets targeting specific perception problems~\cite{burri2016euroc,cordts2016cityscapes,geiger2013vision}. Aerial datasets have enabled advances in large-scale mapping and structure-from-motion methods, while underwater datasets have supported research in image enhancement, segmentation, and detection under degraded visibility conditions. However, datasets that provide synchronized observations of marine environments across aerial, surface, and submerged viewpoints remain limited. In particular, few datasets support both visual tracking and mapping research while also providing localization information that enables quantitative evaluation under real-world environmental conditions~\cite{li2019underwater,pedersen2019detection,wei2024underwater}.

Visual tracking in marine environments is especially challenging due to appearance variation caused by wave motion, lighting changes, and intermittent occlusions. Mapping of shallow-water environments presents additional difficulties as time-varying surface conditions introduce geometric inconsistency and instability. Given the difficulties of collecting underwater data, more datasets are required to expand the publicly available data to all algorithms to be developed and trained to handle these conditions.

This work introduces the \emph{Aerial--Marine Perception Dataset (AMP2026~\footnote{The dataset is publicly available at:
\url{https://huggingface.co/datasets/edwinmeriaux/AMP2026
}}
)}, a multi-platform marine robotics dataset collected during field deployments in both coastal saltwater and freshwater lake environments. AMP2026 is designed to support the evaluation of tracking and mapping across aerial, surface, and submerged viewpoints under realistic marine conditions. The dataset is organized around two primary research categories: tracking and mapping. The tracking sequences contain observations of surface and submerged targets recorded from aerial, surface, and underwater viewpoints under varying motion characteristics and environmental conditions. In approximately half of the tracking sequences, GNSS measurements are available for the tracked platform, enabling quantitative validation of tracking performance without requiring dense manual annotation. The mapping sequences provide repeated geo-referenced observations of the same shallow-water regions across time and viewpoints, supporting evaluation of multi-view map construction, temporal fusion, and distortion-compensation methods.

The primary contributions of this work are as follows:

\begin{itemize}
    \item The \emph{Aerial--Marine Perception Dataset (AMP2026)}, a marine robotics dataset organized into two task categories, \emph{tracking} and \emph{mapping}, as illustrated in Figure~\ref{fig:dataset_overview_blocks}.
    
    \item A collection of tracking sequences capturing surface and submerged targets observed from aerial, surface, and underwater viewpoints, including scenarios with and without GNSS ground truth to support both qualitative and quantitative evaluation of tracking performance.
    
    \item A collection of mapping sequences consisting of repeated, geo-referenced observations of the same shallow-water regions, enabling evaluation of multi-view map construction, temporal consistency, and cross-view alignment under dynamic surface conditions.
    
    \item Data collected across multiple environments and sensing configurations, providing variation in visibility, surface dynamics, and viewpoint while maintaining a consistent dataset organization across both task categories.
\end{itemize}

The remainder of this paper describes the dataset organization, data collection methodology, experimental protocols, and the research tasks supported by the dataset.

The primary objective of this paper is to present the dataset and its collection methodology, enabling future research across these problem domains.

\section{Related Work}

Publicly available datasets have played a central role in advancing perception and autonomy in robotics by enabling standardized evaluation and comparison across methods~\cite{walke2023bridgedata,calli2017yale}. In terrestrial robotics, large-scale datasets combining multiple sensing modalities have enabled rapid progress in object detection, tracking, and mapping tasks. These efforts demonstrate the impact of well-curated, multi-modal datasets on accelerating algorithmic development and benchmarking.

In contrast, marine robotics datasets remain comparatively limited due to the logistical difficulty and cost of data collection, as well as environmental factors such as limited visibility, refraction at the air-water interface, and dynamic illumination that complicate consistent data acquisition. As a result, existing marine datasets typically focus on a single sensing modality or a narrowly defined task, making cross-domain evaluation challenging.

Some prior work has explored multi-modal datasets in outdoor and off-road settings that share partial similarities with marine environments. For example, the German Outdoor and Offroad Dataset (GOOSE) contains 10,000 frames collected from an off-road vehicle equipped with multiple sensing systems, while its GOOSE-X extension introduces semantic annotations~\cite{Hagmanns_2025}. Although not marine-focused, such datasets illustrate the value of multi-sensor integration for perception in challenging outdoor conditions—an element we seek to incorporate in future extensions of our dataset.

Within the maritime domain, aerial datasets have received the most attention to date. The SeaDronesSee dataset~\cite{varga2022seadronessee} serves as a large-scale benchmark for detecting humans and vessels in open water, containing over 54,000 labeled frames and 400,000 instances captured from altitudes ranging between 5 and 260 meters. While highly valuable for Search and Rescue and surface-level perception, it focuses on the detection of targets that remain fully visible above the waterline. In contrast, our dataset addresses the more complex problem of continuous tracking of partially or fully submerged targets, where observations are affected by refraction, surface dynamics, and varying turbidity.

Beyond aerial observation, a substantial body of work has focused on underwater perception datasets aimed at image enhancement, object detection, or semantic segmentation. Datasets such as Enhancement of Underwater Visual Perception (EUVP) provide paired and unpaired underwater imagery designed to improve visual quality through enhancement and restoration methods, addressing color attenuation and low-contrast conditions common in underwater environments~\cite{islam2020fast}. Other datasets emphasize object detection or environmental monitoring, including marine debris detection and biological observation using ROVs or AUVs. These datasets typically consist of labeled still images or short sequences intended for training perception models, rather than supporting long-term tracking or interaction across multiple platforms. Surveys of underwater perception datasets further highlight that most existing resources focus on recognition tasks in relatively controlled or static environments, with limited support for evaluating temporal consistency or cross-view perception~\cite{gonzalez2023survey}.

Recent efforts have also introduced datasets targeting marine autonomy and surface vehicle perception. Multi-modal datasets for autonomous surface vessels provide synchronized sensor data for obstacle detection and classification, supporting navigation and situational awareness research. However, these datasets generally emphasize safety-oriented surface perception tasks and do not include sustained observation of submerged targets or interaction between aerial, surface, and underwater platforms~\cite{nirgudkar2023massmind}.

In parallel, marine mapping datasets have primarily emerged from underwater surveying and photogrammetry applications. Vision-based reconstruction datasets collected using ROVs and AUVs have enabled advances in structure-from-motion, seabed reconstruction, and habitat mapping~\cite{agrafiotis2018underwater}. 

Overall, existing marine robotics datasets tend to emphasize aerial observation, underwater perception, surface autonomy, or mapping in isolation. Datasets that provide synchronized observations across aerial, surface, and submerged viewpoints, while supporting both tracking and mapping evaluation, remain limited. In particular, few datasets provide localization information that enables quantitative evaluation of tracking performance without dense manual annotation, or repeated geo-referenced observations suitable for studying multi-view map construction under time-varying environmental conditions. The dataset presented in this work is designed to address this gap by providing multi-platform observations collected under realistic marine conditions, enabling evaluation of tracking robustness and mapping consistency across sensing modalities and viewpoints.

\section{Dataset Categories}

The proposed AMP2026 dataset is designed to support multiple research directions at the intersection of marine robotics and computer vision. Data were collected during field deployments off the coast of Barbados and in freshwater lakes in Quebec, Canada. Rather than focusing on a single benchmark task, the dataset is organized into two complementary research categories, as illustrated in Figure~\ref{fig:dataset_overview_blocks}.

\begin{figure}[h]
    \centering
    \includegraphics[
        height=\columnwidth,
        trim=0cm 0cm 0cm 0cm,
        clip
    ]{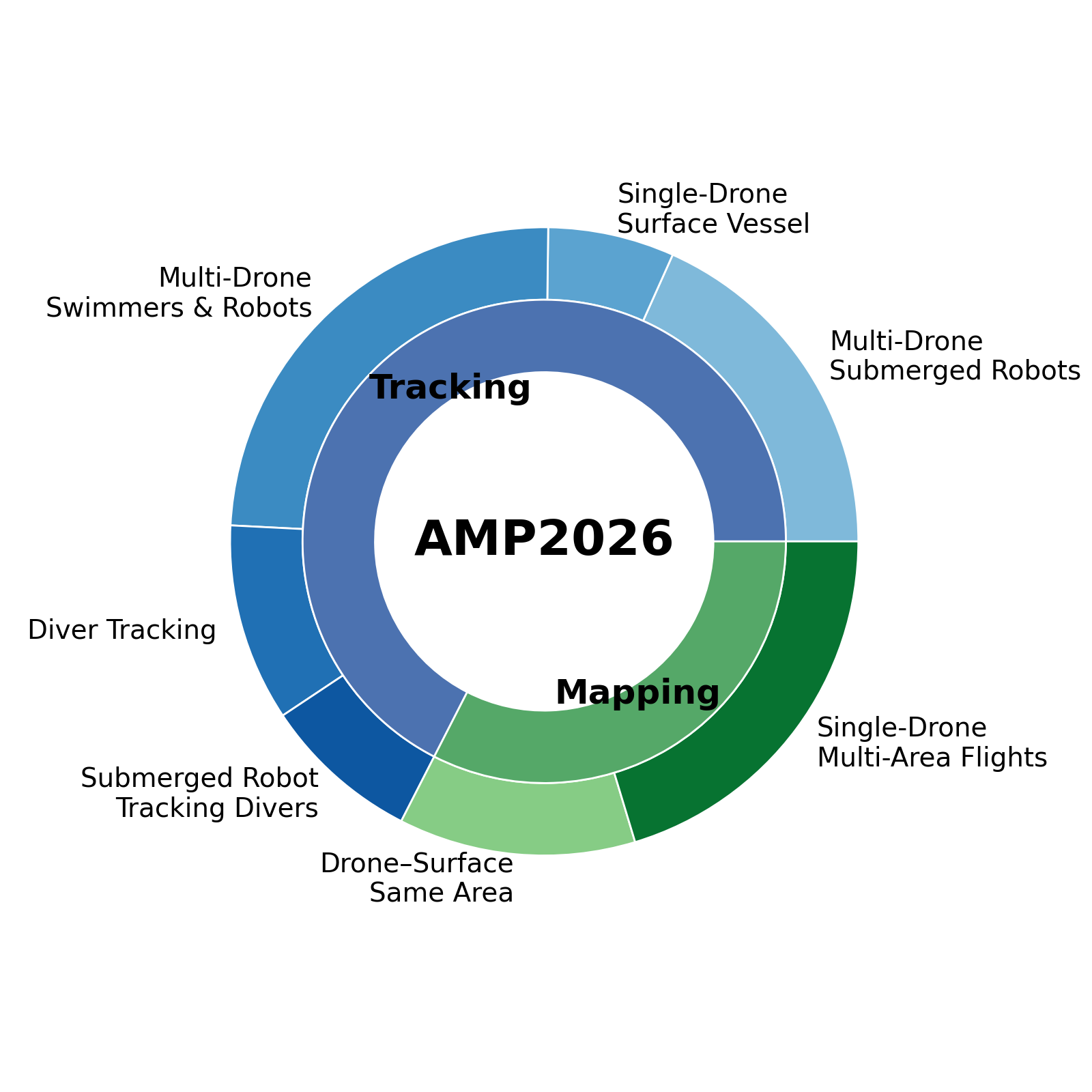}
    \caption{Distribution of data collected in AMP2026 dataset across the different categories and classes in Figure~\ref{fig:dataset_overview_blocks}}
    \label{fig:pie}
\end{figure}

\subsubsection{\textbf{Tracking}}

Visual tracking in marine environments is particularly challenging due to refractive distortions at the air-water interface, surface glint, intermittent occlusions, and motion induced by both sensing platforms and environmental dynamics. Targets observed from aerial, surface, or submerged viewpoints may undergo significant appearance variation caused by wave motion, lighting fluctuations, and partial submergence. These factors complicate robust trajectory estimation and long-term identity preservation, making reliable tracking difficult in real-world marine settings.

This category of the dataset supports research in marine tracking by providing observations of surface and submerged targets captured from multiple sensing platforms. In approximately half of the recorded sequences, GNSS measurements are available for the tracked platform. This enables quantitative validation of visual tracking performance without requiring dense manual annotation and allows evaluation of both visual and positional tracking accuracy, which is often unavailable in maritime tracking datasets. The data collected across different environments enable tracking evaluation under varying environmental conditions, supporting systematic study of robustness to distortion, motion variability, and multi-viewpoint observations. As such, the dataset contributes additional real-world observations to the currently limited corpus of maritime tracking data.

\subsubsection{\textbf{Mapping of Marine Environments}}

Mapping shallow-water and reef environments from purely vision-based robotic observations remains challenging due to water-surface motion, refraction, specular reflections, and lighting variability. Although high-resolution maps can be produced under relatively calm conditions using aerial photogrammetry and structure-from-motion techniques, map construction from time-varying video collected in dynamic marine environments often suffers from geometric inconsistencies and radiometric instability.

The proposed dataset supports research in marine mapping by providing repeated, geo-referenced video observations of the same shallow-water regions across time and viewpoints. This enables systematic evaluation of multi-view map construction, temporal fusion, and distortion-compensation methods under realistic and dynamically varying surface conditions. The dataset includes both aerial and submerged observations of the same spatial regions, supporting investigation of cross-view reconstruction and fusion methods that remain less commonly represented in existing maritime mapping datasets.

\begin{figure}
\centering

\begin{subfigure}[t]{0.48\columnwidth}
    \centering
    \includegraphics[width=\linewidth,
        trim=11cm 0cm 0cm 0cm,
        clip]{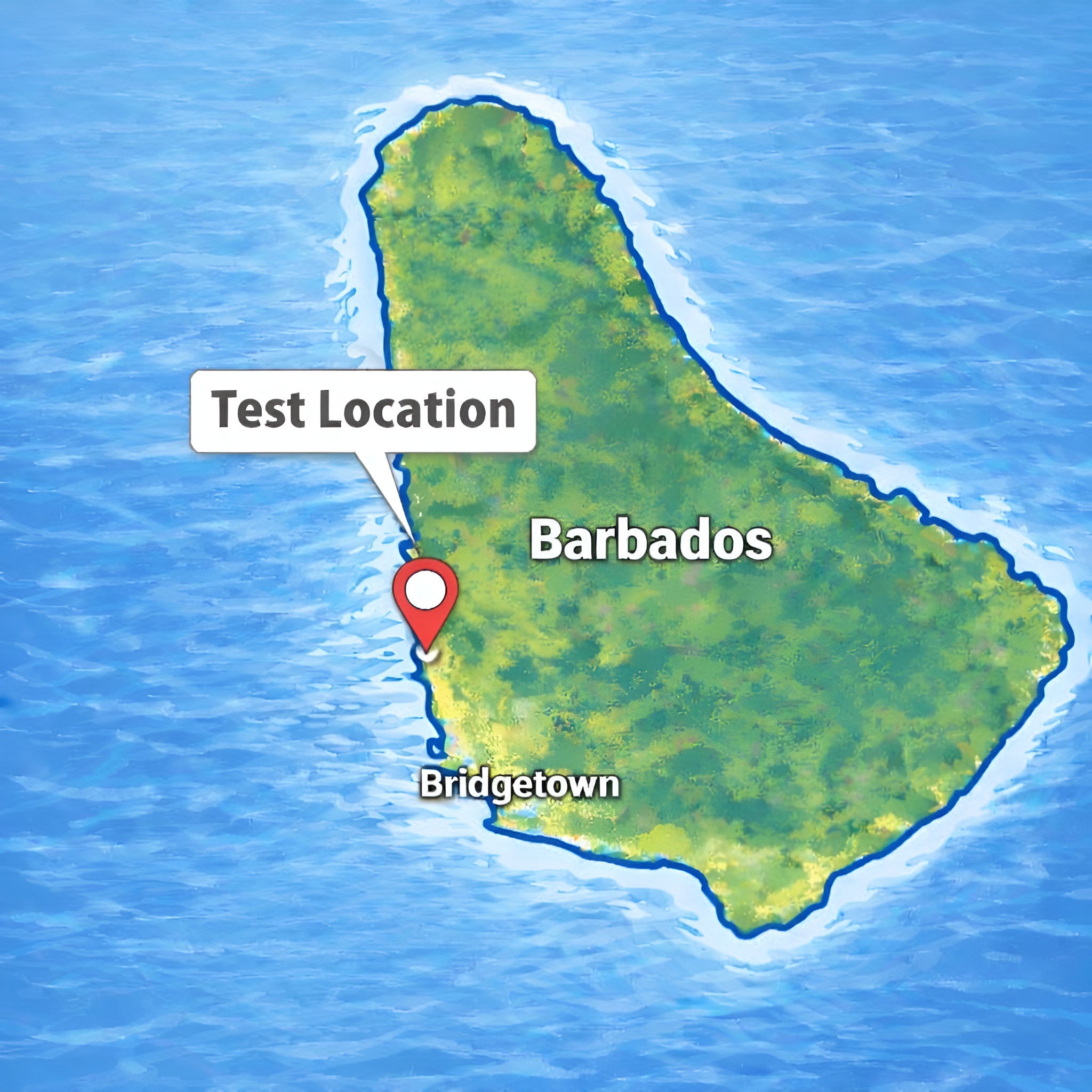}
    \caption{Barbados test location}
    \label{fig:Barbados}
\end{subfigure}
\hfill
\begin{subfigure}[t]{0.48\columnwidth}
    \centering
    \includegraphics[
        width=\linewidth,
        trim=15cm 3cm 12.0cm 1.0cm,
        clip
    ]{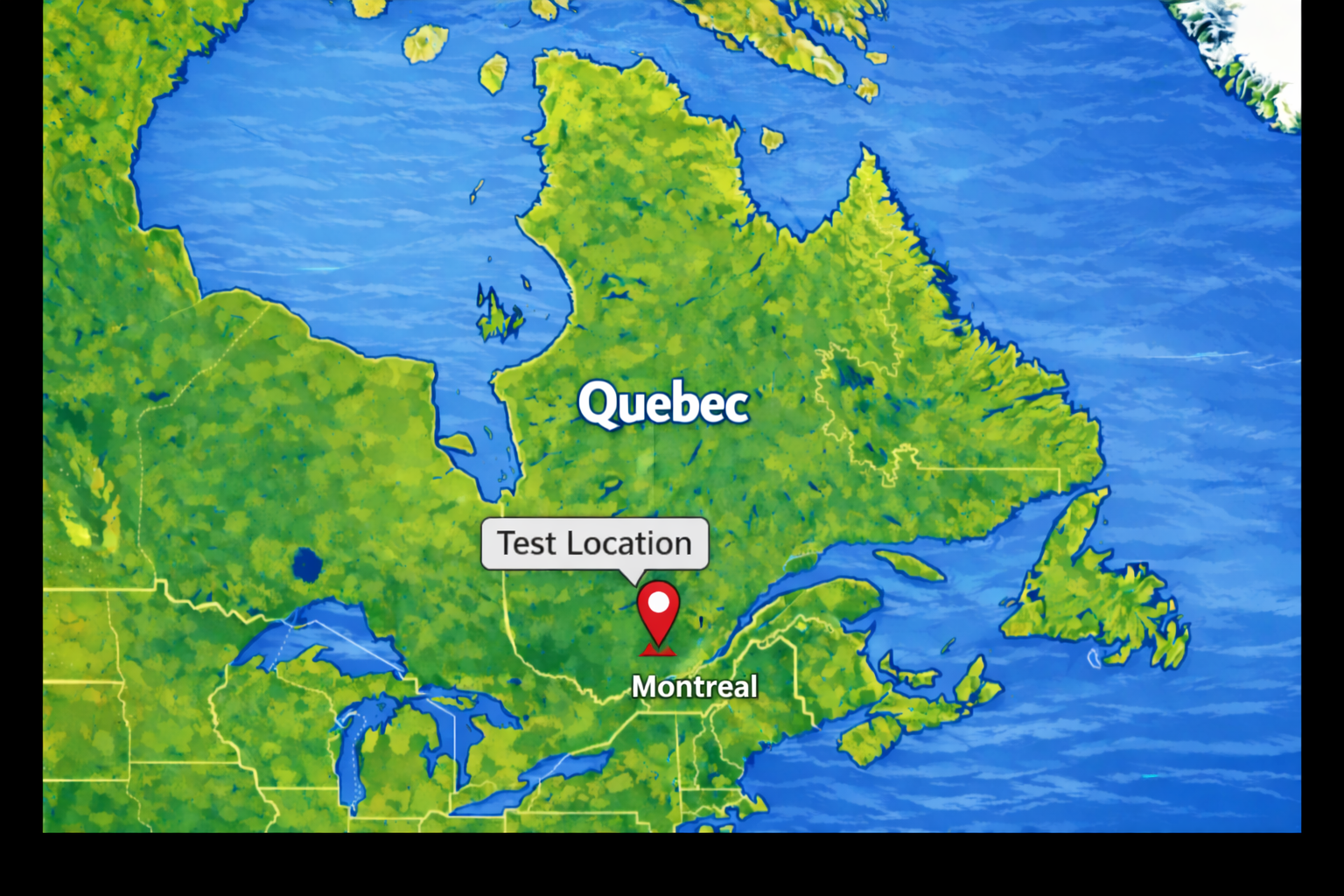}
    \caption{Quebec test location}
    \label{fig:Quebec}
\end{subfigure}

\caption{Dataset collection locations for AMP2026.}
\label{fig:sidebyside}
\end{figure}

\section{Data Collection}

This section describes the environments in which the data were collected and the sensing modalities recorded by each platform. The goal is to clearly document the acquisition setup and establish a standardized data collection framework that can be consistently extended in future deployments. Detailed experimental protocols for each task category are presented in the subsequent section.

\subsection{Collection Sites}

\begin{figure}[t]
    \centering
    \includegraphics[width=\linewidth,
        trim=10.2cm 0cm 12.5cm 0cm,
        clip]{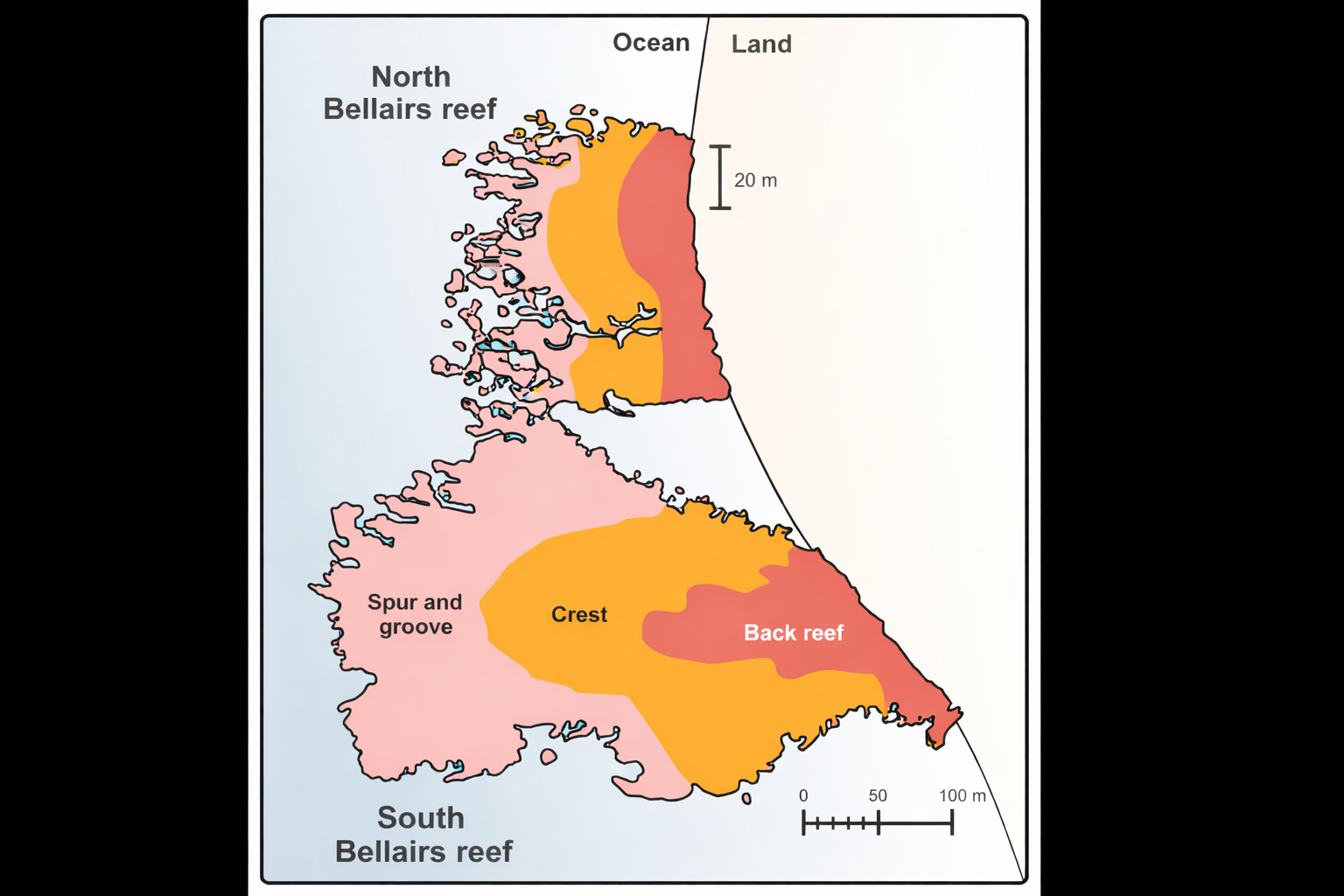}
    \caption{Reefs off the coast of Bellairs Research Institute of McGill University in Barbados. This image was generated with inspiration from the work seen in ~\cite{turgeon2011home}}
    \label{fig:reefs}
\end{figure}

Data were collected during field deployments in two primary locations:

\begin{itemize}
    \item \textbf{Barbados (Saltwater Coastal Reef Environment):} 
    Data were collected off the coast of the Bellairs Research Institute of McGill University in Barbados as seen in Figure~\ref{fig:Barbados}. The study area consists primarily of reefs~\cite{irvine2020barbadosreef} systems composed of both live and dead coral structures\cite{griffith2025tracking}. A sample map of the reefs can be seen in Figure~\ref{fig:reefs}. Experiments were conducted in shallow coastal regions, with robots, human participants, and reef structures of interest located at depths not exceeding ten meters. Water clarity varied across deployments depending on weather conditions, time of day, and sea state. All data were collected under fair-weather conditions.

    \item \textbf{Quebec (Freshwater Lake Environment):}
    Data were collected in Lake Ouareau in southern Quebec, Canada as seen in Figure~\ref{fig:Quebec}. The water exhibited reduced clarity due to sediment and weather-related effects. Tracked objects remained within approximately one meter of the surface, ensuring consistent visibility despite turbidity.
\end{itemize}

These environments provide complementary conditions in terms of water type, visibility, and surface dynamics.

\subsection{Data Collection Platforms}

Across the different task categories, multiple robotic platforms were used depending on the experimental objective. Each platform records visual data along with available localization and inertial measurements.

\subsubsection{\textbf{Aerial Drones}}

\begin{itemize}

    \item Quad-rotor weighing $249$ grams.
    \item Camera: High-resolution RGB camera capturing still images up to 48 MP and video at up to 4K resolution at 30 fps, with an approximate horizontal field of view of 82°. A three-axis stabilized gimbal reduces motion blur during flight.
    
    \item Operating altitude: Flights were conducted between 10 m and 60 m above the water surface, depending on the task and environmental conditions.
    
    \item GNSS and telemetry: Standard GNSS measurements (latitude, longitude, altitude) are recorded along with timestamped platform pose (roll, pitch, yaw) and camera orientation.
\end{itemize}

\subsubsection{\textbf{Submerged Robotic Platform}}

\begin{itemize}

    \item Submersible robot utilizing 6 flippers to navigate.
    \item Camera: Forward-facing synchronized stereo cameras recording video at up to $640 \times 480$ resolution with a frame rate of 10 fps. All frames are timestamped.
    
    \item Inertial data: Orientation (quaternion), angular velocity, and linear acceleration are recorded and timestamped using a common clock shared with the camera to facilitate temporal alignment.
\end{itemize}

\subsubsection{\textbf{Diver-Based Data Collection}}

\begin{itemize}
    \item Camera: Compact waterproof action camera capturing 4K video at a minimum of 30 fps with a $155^\circ$ field of view.
    
    \item Recording conditions: The camera was handheld by divers and swimmers operating at shallow depths, either directly or mounted on a handheld pole to vary viewpoint and capture nearby submerged structures and robotic platforms.
\end{itemize}

\subsubsection{\textbf{Surface Vessel}}

\begin{itemize}
    \item Propeller-based surface water vehicle.
    \item Camera: A downward-facing RGB camera mounted beneath the vessel hull. The camera used is the same as the camera used by the divers and swimmers.
    
    \item GNSS and telemetry: Standard GNSS measurements (latitude, longitude, altitude) are recorded along with timestamped orientation, angular velocity, and linear acceleration.
\end{itemize}


\section{Test Methodology by Category}

In Figure~\ref{fig:dataset_overview_blocks}, each AMP2026 dataset category contains multiple classes of tests designed to capture different sensing configurations and environmental conditions. The distribution of data can be seen in Figure~\ref{fig:pie}. In this section, we describe how each class of test was conducted and specify the data recorded during each experiment.

\subsection{Tracking Sequences}

The tracking component of the dataset consists of five distinct classes of experiments. Each class is designed to evaluate visual tracking performance under different sensing modalities, viewpoints, and motion characteristics. The scenarios range from controlled, repeatable trajectories with ground-truth positioning to unconstrained motion without external localization.

\subsubsection{Multi-Drone Tracking of Submerged Robotic Platforms}

In these sequences, multiple submerged robotic platforms move through the water while maintaining a fixed formation. The robots are towed along predetermined linear paths to ensure repeatable and known trajectories. During each trial, three aerial drones simultaneously observe the submerged robots from distinct viewpoints.

Ground-truth positional information is obtained using GNSS modules mounted on the submerged robots, with antenna elements extending above the water surface. Each aerial drone records synchronized video together with GNSS measurements and full telemetry, including platform pose and camera orientation.

One long 15-minute trial was collected from three different drones' points of view (45 minutes of data total). This test is easily subdivided into each drone's individual data or into partial tests if the video and telemetry data of a given drone are divided into segments. In all cases, the robots followed predefined linear trajectories to enable consistent evaluation of multi-view tracking and cross-view positional estimation.

\subsubsection{Single-Drone Tracking of Surface Vessels}

These sequences consist of a single aerial drone observing an autonomous surface vessel operating in open water. In contrast to the controlled linear trajectories of the previous category, the vessel follows pseudo-random motion patterns designed to introduce heading changes, speed variations, and irregular movement, thereby increasing tracking difficulty.

The aerial drone records video, GNSS measurements, and full telemetry. The surface vessel simultaneously records its own GNSS position and telemetry, enabling direct comparison between visually estimated trajectories and measured ground-truth motion.

Four trials totaling 16 minutes of data were collected. In each trial, the surface vessel departs from the shoreline and traverses open water along non-repetitive paths.

\subsubsection{Multi-Drone Tracking of Swimmers and Submerged Robots}

In these sequences, aerial drones observe swimmers and submerged robotic platforms operating within the same environment. Motion in these trials is task-driven and not predefined, resulting in irregular, non-repetitive trajectories and frequent changes in relative configuration.

The aerial drones record synchronized video, GNSS measurements, and telemetry. The swimmers and submerged robots remain fully submerged during operation and do not provide GNSS-based ground-truth positioning.

Seven trials for each drone were collected, totaling roughly 20 minutes of video data per drone (60 minutes total). Each trial consists of a submerged robot navigating within the environment while two swimmers interact with and maneuver around the robot.

\subsubsection{Single-Diver Tracking of Submerged Robots and Divers}

These sequences are collected from a diver-perspective viewpoint. A diver equipped with a handheld underwater camera records a mission involving a submerged robot and an additional diver. The participants move along predetermined linear paths to provide controlled underwater motion under reduced visibility conditions.

The diver-mounted system records video only and does not provide GNSS or telemetry measurements.

There are 32 trials recording videos totaling 25 minutes of video data. In some cases, both the diver and the aqua robot are in frame. In the other scenarios, only one of the two is in the frame.

\subsubsection{Single-Submerged Robot Tracking of Divers}

In these sequences, a submerged robotic platform follows a diver while recording onboard video and telemetry. The robot maintains visual observation of the diver during motion through the environment, resulting in viewpoint changes driven by underwater navigation dynamics and relative motion between the robot and the diver.

The submerged robot records synchronized stereo video together with onboard telemetry and inertial measurements. No external GNSS ground truth is available due to full submersion of both the sensing platform and the tracked subject.

A total of 32 trials were recorded, consisting of paired video streams and totaling approximately 50 minutes of video data. Depending on relative motion and visibility conditions, some segments contain both the diver and the robotic platform within the field of view, while others contain only one of the two.


\subsection{Mapping}

The mapping component of the dataset consists of two classes of experiments designed to support single-view and multi-view map construction in shallow-water environments. These sequences provide repeated observations of the same geographic areas from aerial and surface viewpoints, enabling comparative map constructions, cross-platform alignment, and multi-view fusion.

\subsubsection{Drone-Surface Vessel Same-Area Observations}

These sequences consist of coordinated observations of the same geographic area from both aerial and surface viewpoints. In each case, an aerial drone first surveys the region in several passes, followed by a surface vessel traversing the same area while recording downward-facing imagery.

The drone records synchronized video, GNSS measurements, and full telemetry, including platform pose and camera orientation. The surface vessel records onboard video together with GNSS and telemetry data. This configuration provides two complementary perspectives of the same environment: a wide-area aerial view and a lower-altitude, surface-proximal view.

A total of ten video sequences were collected for this category. Three of these sequences include the surface vessel within the drone’s field of view, enabling improved temporal alignment and cross-platform synchronization. Each video sequence is several minutes in duration.

These experiments support research in cross-view map construction, multi-platform spatial alignment, and multi-modal map construction of shallow-water regions.

\subsubsection{Single-Drone Multi-Area Flights}

In these sequences, a single aerial drone performs repeated flights over defined coastal and shallow-water areas. Each area is surveyed multiple times under varying environmental conditions, including changes in lighting, surface disturbance, and water clarity. The repeated coverage provides temporal redundancy and viewpoint variation for map construction and image fusion.

During each flight, the drone records high-resolution video together with synchronized GNSS measurements and full telemetry, including platform pose and camera orientation. Flight paths are designed to provide sufficient spatial overlap between passes to support structure-from-motion, orthomosaic generation, and multi-view map construction.

Twenty video sequences of varying duration were collected in this category, each with associated GNSS data. Multiple geographic areas were surveyed, with repeated flights conducted over each location to enable evaluation of map consistency, temporal stability, and environmental variability. 

Some of these sequences overlap spatially with the Drone-Surface Vessel experiments, allowing cross-category reuse for multi-platform alignment studies.

\section{Sample Dataset Usage}

\begin{figure}[h]
    \centering
    \includegraphics[width=0.9\linewidth,
        trim=0cm 0cm 0cm 0cm,
        clip]{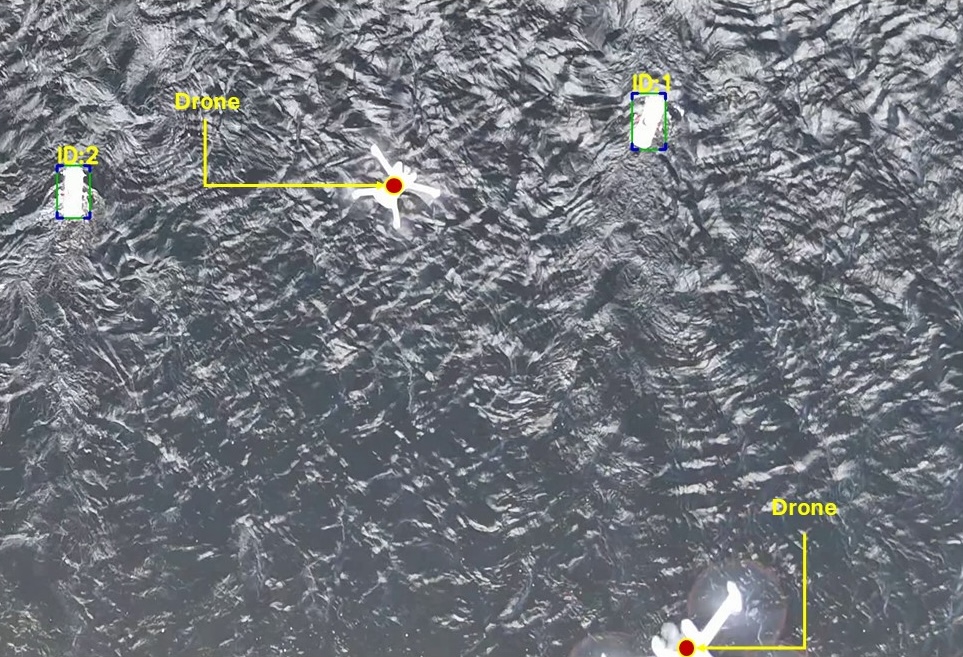}
    \caption{Tracking Category 1: Top-down aerial image sample displaying a multi-drone tracking system in action to track two submerged vehicles (ID:1 and ID:2). }
    \label{fig:tracked_aquas}
\end{figure}

In this section, we present example use cases of the AMP2026 dataset for both tracking and mapping tasks. The dataset supports multi-drone tracking of submerged robotic platforms, where multiple aerial viewpoints enable observation of submerged and surface targets simultaneously. Figure~\ref{fig:tracked_aquas} illustrates a representative example of top-down aerial tracking from the multi-drone tracking sequences, showing two submerged robotic vehicles observed from multiple viewpoints. In these experiments, GNSS ground truth is provided for the robotic platforms, enabling quantitative evaluation of tracking performance.

\begin{figure}[h]
    \centering
    \includegraphics[width=0.9\linewidth,
        trim=0cm 0cm 0cm 0cm,
        clip]{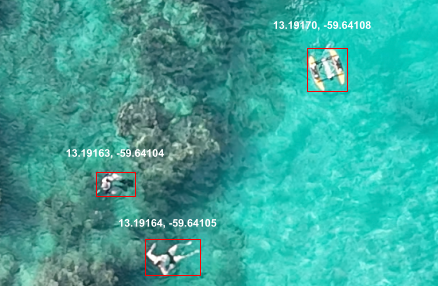}
    \caption{Tracking Category 2: Sample aerial footage of surface vessel with estimated GPS coordinates }
    \label{fig:robot_tracked_divers}
\end{figure}

The dataset also supports single-drone tracking of surface vessels operating in open water. In this scenario, aerial footage captures the motion of a surface vehicle while estimated GPS coordinates are overlaid on the imagery, as shown in Figure~\ref{fig:robot_tracked_divers}. This configuration enables comparison between visually estimated trajectories and measured positional data.

\begin{figure}[h]
    \centering
    \includegraphics[width=0.9\linewidth,
        trim=0cm 0cm 0cm 0cm,
        clip]{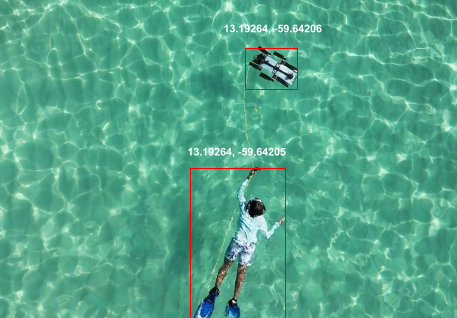}
    \caption{Tracking Category 3: Sample aerial footage of submerged vehicle and swimmer with estimated GPS coordinates }
    \label{fig:cat3}
\end{figure}

In addition, the dataset contains multi-drone tracking sequences involving swimmers and submerged robotic platforms observed simultaneously from aerial viewpoints. These sequences provide challenging tracking scenarios characterized by irregular motion and appearance variation, allowing both submerged and surface agents to be tracked concurrently, as illustrated in Figure~\ref{fig:cat3}.

\begin{figure}[h]
    \centering
    \includegraphics[width=\linewidth,
        trim=1.1cm 2.1cm 2.5cm 1.5cm,
        clip]{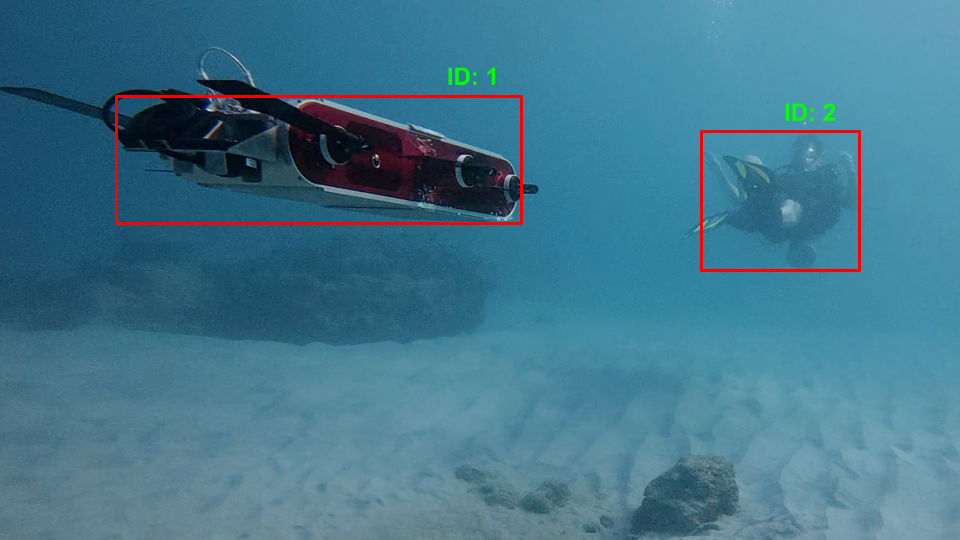}
    \caption{Tracking Category 4: Sample underwater footage from a diver-mounted camera utilizing YOLO detection to track fellow divers and a submersible robot. The objects are being tracked with their bounding boxes shown.}
    \label{fig:aqua}
\end{figure}

\begin{figure}[h]
    \centering
    \includegraphics[width=0.9\linewidth,
        trim=0cm 0cm 0cm 0cm,
        clip]{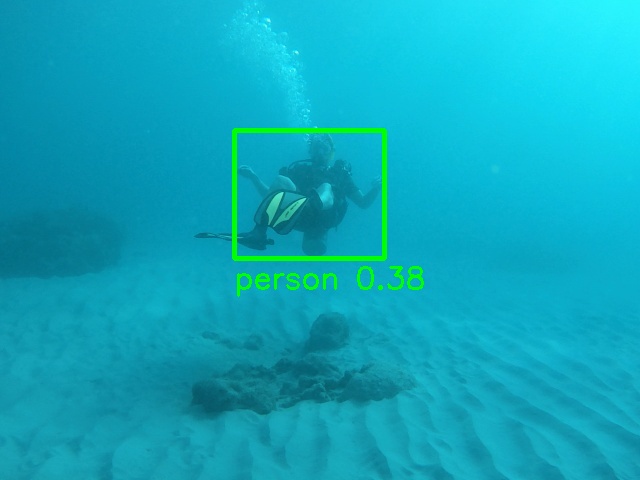}
    \caption{Tracking Category 5: Sample underwater footage from the left camera of the underwater vehicle utilizing YOLO detection to track divers. The diver is identified as category 'person' with confidence 0.38. }
    \label{fig:cat5}
\end{figure}

Beyond aerial tracking, the dataset supports underwater object detection and tracking research. The recorded imagery includes challenging visual conditions such as turbidity, lighting variability, and partial occlusions. Figure~\ref {fig:aqua} presents a sample tracked case of a submerged robot and diver from the point of view of another diver, while Figure \ref{fig:cat5} presents sample YOLO~\cite{yolo26_ultralytics} detection results from the robot-mounted cameras from the same trial.

\begin{figure}[t]
    \centering
    \includegraphics[
        width=0.9\linewidth,
        trim=1.1cm 2.1cm 2.5cm 1.5cm,
        clip
    ]{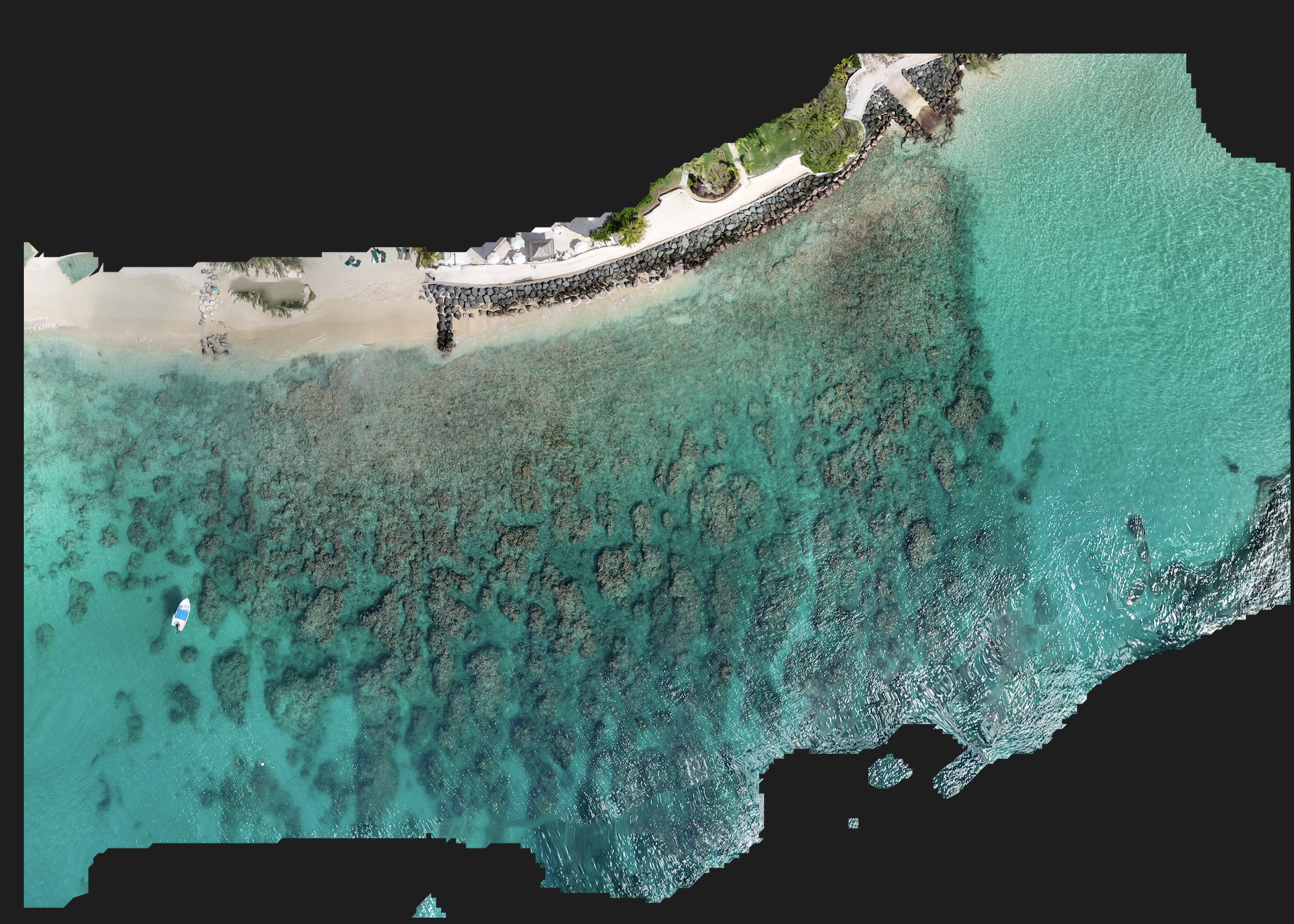}
    \caption{Sample composite map of Northern Bellairs reef from Figure~\ref{fig:reefs}.}
    \label{fig:NB}
\end{figure}

\begin{figure}[t]
    \centering
    \includegraphics[
        width=0.9\linewidth,
        trim=2cm 1cm 0cm 0cm,
        clip
    ]{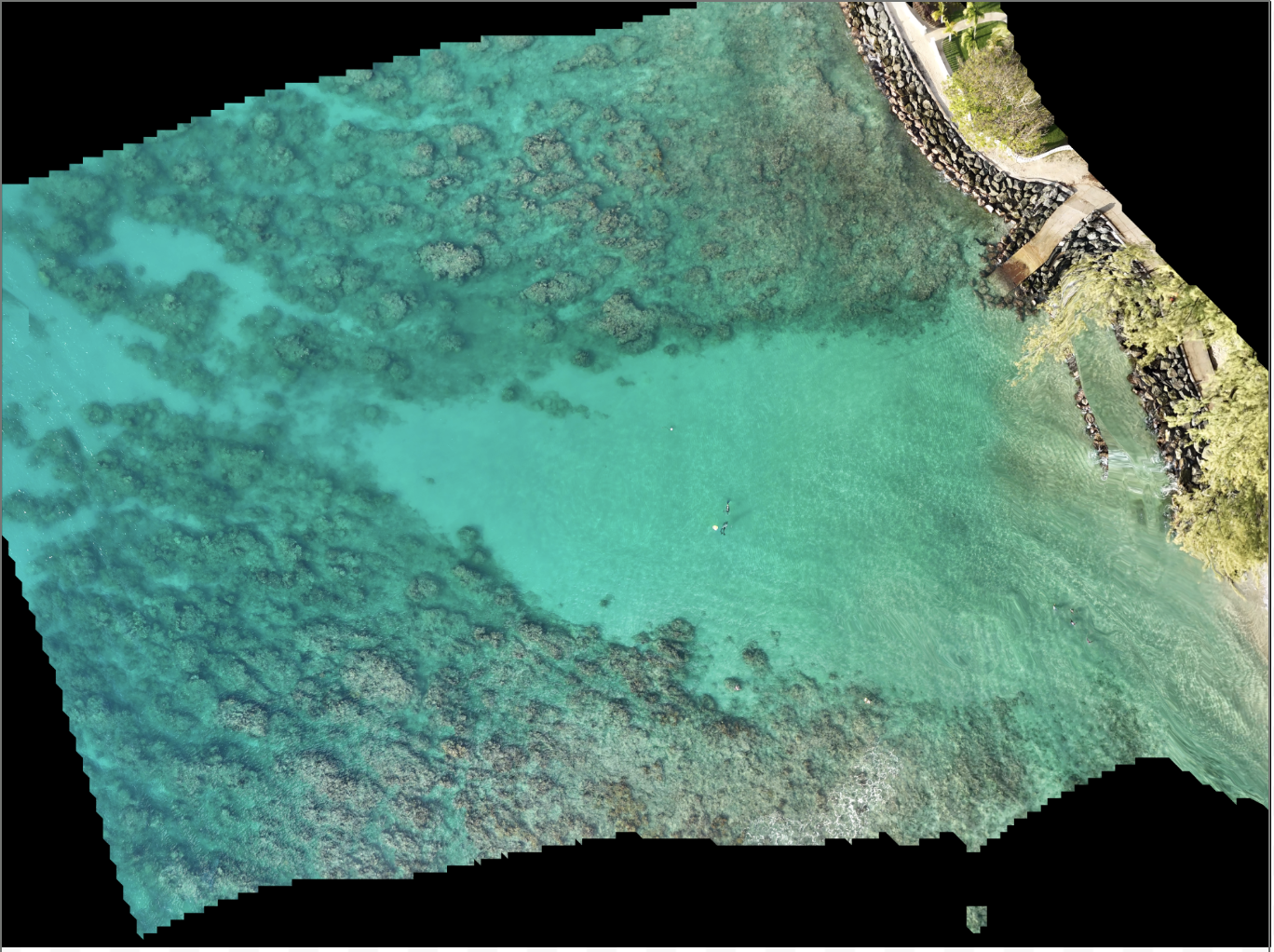}
    \caption{Sample composite map between North and South Bellairs from Figure~\ref{fig:reefs}.}
    \label{fig:B}
\end{figure}

Finally, the mapping sequences enable reconstruction of shallow-water environments from aerial and surface observations. Figures~\ref{fig:NB} and \ref{fig:B} present sample composite maps generated from the North and South Bellairs reef data using OpenDroneMap~\cite{vacca2020web}. These examples highlight the dataset’s suitability for generating orthomosaics and structural maps of coral reef environments, as well as for evaluating multi-view map construction and temporal consistency under dynamic surface conditions.

\vspace{-10pt}

\section{Conclusion}

This paper presented the \emph{Aerial--Marine Perception Dataset (AMP2026)}, a multi-platform marine robotics dataset designed to support research in visual tracking and marine environment mapping under realistic environmental conditions. The dataset provides synchronized observations from aerial, surface, and submerged platforms collected across both coastal and freshwater environments, capturing challenges arising from refraction, dynamic surfaces, turbidity, and viewpoint variability. By organizing the data into tracking and mapping categories and providing localization information where available, AMP2026 enables systematic evaluation of perception algorithms in settings that remain underrepresented in existing datasets.

AMP2026 is intended to serve as a resource for future research in marine perception, multi-view tracking, and vision-based mapping across heterogeneous robotic platforms. Future work will focus on expanding the dataset through additional deployments, increasing environmental diversity, and incorporating semantic annotations and evaluation benchmarks to further support standardized comparison of tracking and mapping methods.

\section{Acknowledgments}

The authors would like to acknowledge all participants of the 2026 Annual Marine Robotics Workshop and Field Robotics Trials in Barbados for their assistance in the planning and coordination required for data collection. The authors also thank Jeremy Mallette from Independent Robotics for his support in the deployment of robotic platforms. We further acknowledge Sasha Nicolas Dolgopolyy and Charles Benjamin from the Mobile Robotics Lab (MRL) for their assistance in preparation for the data collection.

During the writing and editing of this work, the authors used ChatGPT to have assistance in the generation of certain figures and grammar editing. After this tool was used, the authors reviewed, edited, and confirmed the correctness of the content generated. 

\bibliographystyle{IEEEtran}
\bibliography{Biblio}

\end{document}